\RequirePackage{xcolor}
\documentclass{article}
\usepackage[final]{corl_2019}
\usepackage[utf8]{inputenc}
\usepackage{cite}
\usepackage{graphicx}
\usepackage{caption}
\usepackage{subcaption}
\usepackage{url}
\usepackage{amssymb}
\usepackage{amsmath}
\usepackage{algorithm}
\usepackage{algorithmic}
\usepackage{wrapfig}

\graphicspath{ {Figures/} }

\usepackage[textsize=tiny]{todonotes}

\newcommand{\curiousilqr}{\textsl{curious iLQR }}

\newcommand{\BR}[1]{\boldsymbol{\mathrm{#1}}}

\title{Curious iLQR: Resolving Uncertainty in Model-based RL}

\author{Sarah Bechtle$^{1,2}$, Yixin Lin$^{2}$, Akshara Rai$^{2}$, Ludovic Righetti$^{1,3}$ and Franziska Meier$^{2}$\\
$^{1}$Max Planck Institute for Intelligent Systems\\
$^{2}$Facebook AI Research, $^{3}$New York University\\
\texttt{\{sbechtle,ludovic.righetti\}@tuebingen.mpg.de}\\
\texttt{\{yixinlin,akshararai,fmeier\}@fb.com}
}

\begin{document}
\maketitle

\begin{abstract}
Curiosity as a means to explore during reinforcement learning problems has recently become very popular. However, very little progress has been made in utilizing curiosity for learning control. In this work, we propose a model-based reinforcement learning (MBRL) framework that combines Bayesian modeling of the system dynamics with \curiousilqr, an iterative LQR approach that considers model uncertainty. During trajectory optimization the \curiousilqr attempts to minimize both the task-dependent cost and the uncertainty in the dynamics model. We demonstrate the approach on reaching tasks with 7-DoF manipulators in simulation and on a real robot. Our experiments show that MBRL with \curiousilqr reaches desired end-effector targets more reliably and with less system rollouts when learning a new task from scratch, and that the learned model generalizes better to new reaching tasks.
\end{abstract}
%
\section{Introduction}

Model-based reinforcement learning holds promise for sample-efficient learning on real robots \citep{atkeson1997comparison}. The hope is that a model learned on a set of tasks can be used to learn to achieve new tasks faster. A challenge is then to ensure that the learned model generalizes beyond the specific tasks used to learn it. We believe that curiosity, as means of exploration, can help with this challenge. Though curiosity has been defined in various ways, it is generally considered a fundamental building block of human behaviour \citep{loewenstein1994psychology} and essential for the development of autonomous behaviour \citep{white1959motivation}. 

In this work, we take inspiration from \citep{Kagan1972}, which defines curiosity as motivation to resolve uncertainty in the environment. Following this definition,
we postulate that by seeking out uncertainties, a robot is able to learn a model faster and therefore achieve lower costs more quickly compared to a non-curious robot. Keeping real robot experiments in mind, our goal is to develop a model-based reinforcement learning (MBRL) algorithm that optimizes action sequences to not only minimize a task cost but also to reduce model uncertainty.

\begin{wrapfigure}{o}{0.5\textwidth}
\vspace{-1.2cm}
\begin{center}
    \includegraphics[width=0.5\textwidth, trim={4.0cm 6.0cm 4.0cm 4.0cm}, clip]{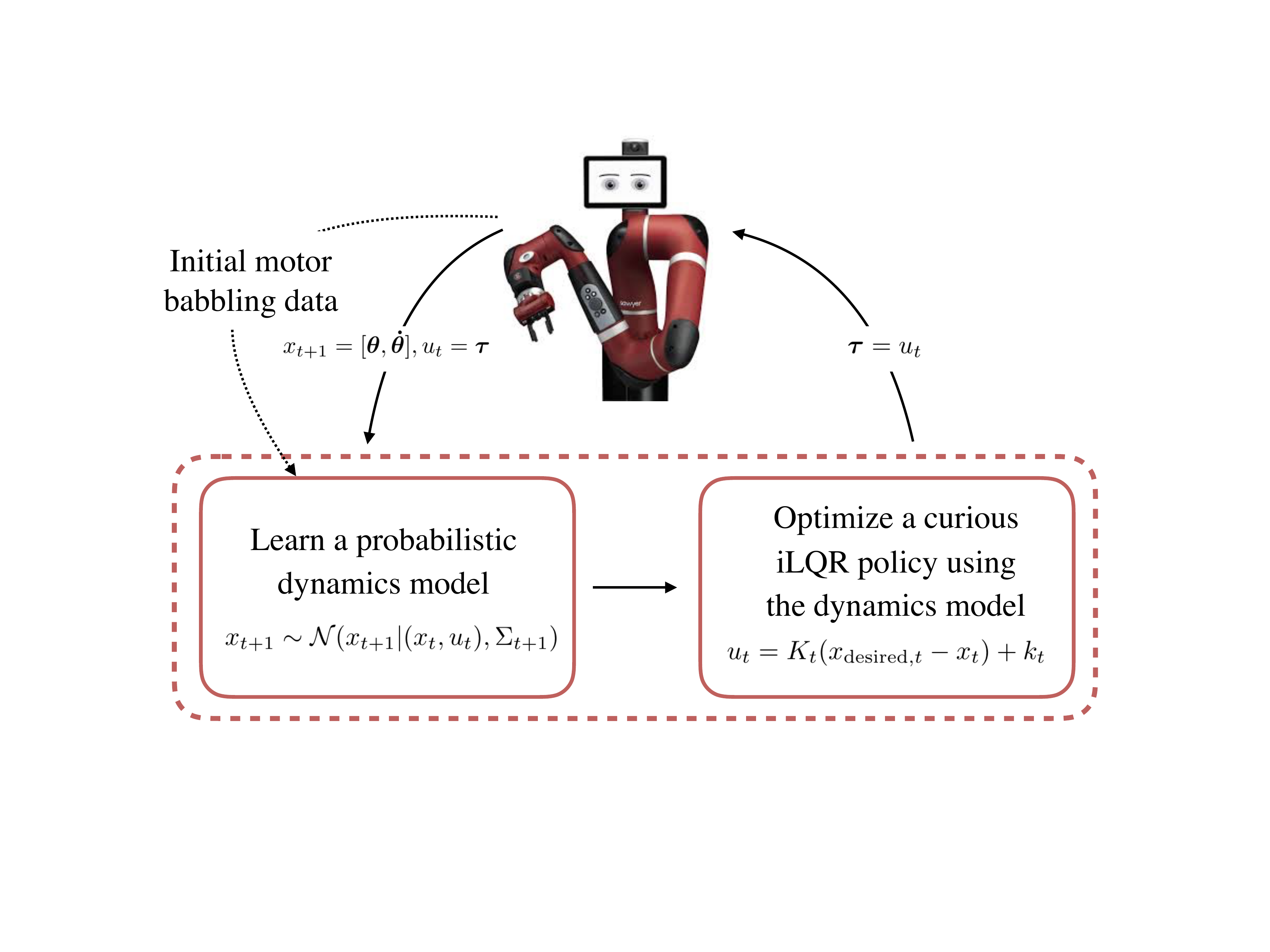}
    \vspace{-0.7cm}
    \caption{\small{Approach overview: motor babbling data initializes the dynamic model, the main loop then alternates between model learning and policy updates.}}
    \label{fig:overview}
\end{center}
\vspace{-1cm}
\end{wrapfigure}
Specifically, our MBRL algorithm iterates between learning a probabilistic model of the robot dynamics and using that model to optimize local control policies (i.e. desired joint trajectories and feedback gains) via a \textsl{curious} version of the iterative Linear Quadratic Regulator (iLQR) \citep{Tassa2014}. These policies are executed on the robot to gather new data to improve the dynamics model, closing the loop, as summarized in Figure~\ref{fig:overview}. 

In a nutshell, our curious iLQR aims at optimizing local policies that minimize the cost \textsl{and} explore parts of the model with high uncertainty.
In order to encourage actions that explore states for which the dynamics model is uncertain, we incorporate the variance of the model predictions into the cost function evaluation.
We propose a computationally efficient approach to incorporate this uncertainty by leveraging results on risk-sensitive optimal control \citep{Jacobson1973, Farshidian}.
\citep{Jacobson1973} showed that optimizing actions with respect to the expected exponentiated cost directly takes into account higher order moments of the cost distribution while affording the explicit computation of the optimal control through Riccati equations. A risk-sensitive version of iLQR was recently proposed in \citep{Farshidian}. While in these approaches the dynamic model is typically  considered known and uncertainty comes from external disturbances, we propose to instead explicitly incorporate model uncertainty in the algorithm to favor the exploration of uncertain parts of the model.
The proposed coupling between model learning and risk-sensitive control explicitly favours actions that resolve the uncertainty in the model while minimizing a task-related cost.

The contributions of this work are as follows: 1) We present a MBRL algorithm that learns a global probabilistic model of the dynamics of the robot from data and show how to utilize the uncertainty of the model for exploration through our curious iLQR optimization. 2) We demonstrate that our MBRL algorithm can scale to seven degree of freedom (DoF) manipulation platform in the real world without requiring demonstrations to initialize the MBRL loop. 3) The results show that using curiosity not only learns a better model faster on the initial task, but also that this model generalizes to new tasks more reliably. We perform an extensive evaluation in both simulation and on hardware.
%
\section{Background}\label{section.background}
The goal of MBRL is to solve a task through learning a model $f$ of the true dynamics $f_\text{real}$ of the system that is subsequently used to solve an optimal control problem. The dynamics are described through $x_{t+1} = f(x_t,u_t)$ where $x_t$ and $u_t$ are the state and action of the current time step, and $x_{t+1}$ the state at the next time step. $f$ represents the learned model of the dynamics. MBRL seeks to find a policy $u_t = \pi(x_t)$ that minimizes a cost $\mathcal{J}(x_t,u_t)$ describing the desired behavior. Policy optimization can be performed in various ways such as trajectory sampling approaches as summarized and evaluated in \citep{chua2018deep}, random shooting methods, where trajectories are randomly chosen and evaluated with the learned model, or iterative LQG approaches, as in \citep{levine2013guided}. Model learning also can be tackled with various methods. \citep{levine2014learning} proposes learning linear models of the forward dynamics. In \citep{chua2018deep} the dynamics are learned with an ensemble of neural networks. In general, the learned model of dynamics can be deterministic as in \citep{levine2014learning} or probabilistic as in \citep{deisenroth2011pilco,chua2018deep}.

In MBRL, the learned model is used to simulate the robot behaviour when optimizing a trajectory or control policy. The learned model and the optimizer are task independent; this independence promises sample efficiency and generalization capabilities, as an already learned model can be reused for new tasks. As a side effect, however, the learned models quality can drastically affect the computed solution, as pointed out in \citep{schaal1997learning,atkeson1997comparison,deisenroth2010efficient}, since the policy is optimized given the current learned model and not by interacting with the robot. This effect is called model bias \citep{deisenroth2010efficient} and can lead to a policy with drastically lower performance on the real robot. We argue that exploration can alleviate this model-bias. Resolving model uncertainty while optimizing for a task can encourage visiting states which resolve ambiguities in the learned model and therefore lead to both better models and control policies.

\subsection{Intrinsic motivation for RL}
The concept of curiosity has also been explored within the reinforcement learning literature from various angles. For example, a first attempt towards intrinsically motivated agents consisted in rewarding agents to minimize prediction errors of sensory events \citep{Barto2004, Singh2004, Singh2010}. This initial work was designed for low-dimensional and discrete state-and-action spaces. Recently, curiosity as a means to better explore was also investigated for high-dimensional continuous state spaces \citep{Bellemare2016, Pathak2017}. Most of this work, including recent efforts towards curiosity driven robot learning \citep{tanneberg2019intrinsic, laversanne2018curiosity}, has defined curiosity as a function of model prediction error and within a model-free reinforcement learning framework. In MBRL, \citep{shyammax} recently proposed a measure of disagreement as exploration signal. \citep{levine2016end} propose a maximum entropy exploration behaviour. Other algorithms which take uncertainty into account have been presented as well \citep{deisenroth2011pilco,williams2017information,chua2018deep,boedecker2014approximate}. They differ in their choice of policy optimization, dynamics model representation and how they incorporate uncertainty. While \citep{deisenroth2011pilco, chua2018deep} utilize model uncertainty to generate trajectory distributions, the uncertainty does not play an explicit role in the cost. Thus, these approaches do not explicitly optimize actions that resolve uncertainty in the current model of the dynamics, which is in contrast to the approach we propose in this paper.

\subsection{Risk Sensitive stochastic optimal control}\label{sec.risk_sensitive_optimal_control}
Risk-sensitive optimal control has a long history \citep{Jacobson1973, WHITTLE:1981cd}.
The central idea is to not only minimize the expectation
of the performance objective under the stochastic dynamics but to also take into account higher-order moments of the cost distribution.
The objective function takes the form of an exponential transformation of the performance criteria $ J=\min _{ \pi  }{\mathbb{E}\left\{\mathrm{exp}[\sigma \mathcal{J}  \left( \pi  \right)  \right]\} }$ \citep{Jacobson1973}.
Here, $\mathcal{J}(\pi)$ is the performance index, which is a random variable, and a functional of the policy $\pi$. $\mathbb{E}$ is the expected value of $\mathcal{J}$ over stochastic trajectories induced by the policy $\pi$.
  $\sigma \in \mathbb{R}$ accounts for the sensitivity of the cost to higher order moments (variance, skewness, etc.). 
  Notably, from \citep{Farshidian}, the cost is $\frac{1}{\sigma} \log(J)  = \mathbb{E}(\mathcal{J}^*)+ \frac{\sigma}{2} \mathrm{var}(\mathcal{J}^*) + \frac{\sigma^2}{6} \mathrm{sk}(\mathcal{J}^*) + \cdots$,
where $\mathrm{var}$ and $\mathrm{sk}$ stand for variance and skewness and $\mathcal{J}^*$ is the optimal task cost.
  When $\sigma>0$ the optimal control will be risk-averse, favoring low costs with low variance but
  when $\sigma<0$ the optimal control will be risk-seeking, favoring low costs with high variance. 
  $\sigma =0$, reduces to the standard, risk-neutral, optimal control problem.
Jacobson \citep{Jacobson1973} originally demonstrated that for linear dynamics and quadratic costs the optimal control could be computed as the solution of a Riccati equation.
Leveraging this result, \citep{Farshidian} recently proposed a risk-sensitive extension of iLQR and \citep{Ponton} further extended the approach to explicitly incorporate measurement noise. 
\section{MBRL via Curious iLQR}\label{section.curious-ilqr}
We present our approach to incorporate curious behaviour into a robot's learning control loop. We are interested in minimizing a performance objective $\mathcal{J}$ to achieve a desired robot behavior and approximate the true dynamics of the system with a discrete-time dynamical system
\begin{equation}
    \boldsymbol{\mathrm{x}}_{t+1}=\boldsymbol{\mathrm{x}}_t+\boldsymbol{\mathrm{f}}\left( \boldsymbol{\mathrm{x}}_t,\boldsymbol{\mathrm{u}}_t \right) \Delta t 
\end{equation}
where $\BR{x}_t$ denotes the state of the system at time step $t$ and $\boldsymbol{\mathrm{f}}$ represents the unknown model of the dynamics of the system and needs to be learned to achieve the desired task. The hypothesis we seek to confirm is that, by trying to explore uncertain parts of the model, our MBRL algorithm can learn a good dynamics model more quickly and find behaviors with higher performance.
Our algorithm learns a probabilistic model of the system dynamics while concurrently optimizing a desired cost objective (Figure \ref{fig:overview}).
It combines i) a risk-seeking iLQR algorithm and ii) a probabilistic model of the dynamics.
We describe the algorithm in the following.
In particular, we show how to incorporate model uncertainty in risk-sensitive optimal control. 
Algorithm \ref{algo:mbrl_loop} shows the complete algorithm.

\begin{minipage}{0.41\textwidth}
\vspace{-0.4cm}
\begin{algorithm}[H]
\small{
\begin{algorithmic}[1]
\STATE{$\mathcal{D} \gets \text{motor babbling data}$}
\STATE{$ \text{train model } f \text{on } D$}
\WHILE{$i < \text{iter}$}
\STATE{$\pi \gets \text{optimize policy via \text{Alg 3}}$}
\STATE{$D_\text{new} \gets \text{rollout $\pi$ on system}$}
\STATE{$\mathcal{D} =\mathcal{D} \cup D_\text{new}$}
\STATE{$\text{train model } f \text{on } \mathcal{D}$}
\ENDWHILE
\end{algorithmic}
\caption{MBRL Algorithm}
\label{algo:mbrl_loop}
}
\end{algorithm}
\vspace{-0.5cm}
\begin{algorithm}[H]
\small{
\begin{algorithmic}[1]
\STATE{$x^\text{new}_0 \gets x_0$}
\WHILE{$t < T$}
\STATE{$\tau^\text{new}_t \gets \tau_t + \alpha k_t + K_t (x_t - x^\text{new}_t)$}
\STATE{$x^\text{new}_{t+1} \gets f(x^\text{new}_t, \tau_t$)}
\ENDWHILE
\STATE{return $\tau^\text{new}, x^\text{new}$}
\end{algorithmic}
\caption{\small{simulate-policy($x,\tau, k, K,\alpha)$}}
}
\end{algorithm}
\end{minipage}
\hfill
\begin{minipage}{0.56\textwidth}
\vspace{-0.4cm}
\begin{algorithm}[H]
\small{
\begin{algorithmic}[1]
\STATE{$\tau \gets \text{Initial random torque trajectory}$}
\STATE{$x^{*} \gets \text{unroll $\tau$ using $f$}$}
\STATE{$\mathrm{\mathbb{A}} \gets \text{Line search parameters, [0,\dots,1]}$}
\STATE{$J^* \gets \text{Optimal iLQR cost so far}$}
\WHILE{$i<\text{opt iter}$}
\STATE{$k,K \gets \text{backward pass, see \ref{sec.curious_recursions}}$} 
\FOR{ $\alpha \in \mathrm{\mathbb{A}}$} 
\STATE{$\tau^{new}, x^{new} \gets \text{simulate-policy}(x, \tau, k, K, \alpha)$}
\STATE{$J_{new} \gets \text{Compute cost of $\tau^\text{new}, x^\text{new}$}$}
\IF{$J_{new}<J^*$}
\STATE{$\tau, x^{*} \gets x^\text{new}, \tau^{new}$}
\ENDIF
\IF{converged}
\STATE{return $\pi(x) = \tau + K (x - x^{*})$}
\ENDIF
\ENDFOR
\ENDWHILE
\end{algorithmic}
\caption{curious-iLQR}
\label{algo:curious-iLQR}
}
\end{algorithm}
\end{minipage}

\subsection{Risk-sensitive iLQR}\label{sec.risk_sensitive}
Consider the following general nonlinear stochastic difference equation
\begin{equation}\label{eq.dynamics}
    \boldsymbol{\mathrm{x}}_{t+1}=\boldsymbol{\mathrm{x}}_t+\boldsymbol{\mathrm{f}}\left( \boldsymbol{\mathrm{x}}_t,\boldsymbol{\mathrm{u}}_t \right) \Delta t +\boldsymbol{\mathrm{g}}\left( \boldsymbol{\mathrm{x}}_t,\boldsymbol{\mathrm{u}}_t\right) \Delta \omega 
\end{equation}
where $\boldsymbol{\mathrm{g}}$ maps a Brownian motion $\Delta \omega$, with 0 mean and covariance $(\boldsymbol{\Sigma} \cdot \Delta t)$, to system states. $\Delta \omega$ and the nonlinear map $\mathbf{g}$, 
typically model an unknown physical disturbance, while assuming a known model $\boldsymbol{\mathrm{f}}$ of the dynamics. 
When considering the exponentiated performance criteria $ J=\min _{ \pi  }{\mathbb{E}\left\{\mathrm{exp}[\sigma \mathcal{J}  \left( \pi  \right)  \right]\} }$ (see \ref{section.background} for more details), it has been shown that iLQR \citep{Tassa2014} can be extended to risk-sensitive stochastic nonlinear optimal control problems \citep{Farshidian}. The algorithm begins with a nominal state and control input trajectory $\boldsymbol{\mathrm{x^{n}}}$ and $\boldsymbol{\mathrm{u^{n}}}$. 
The dynamics and cost are approximated to first and second order respectively along the nominal trajectories $\boldsymbol{\mathrm{u^{n}_{t}}}$, $\boldsymbol{\mathrm{x^{n}_{t}}}$ in terms of state and control deviations
$\delta \boldsymbol{\mathrm{x_{t}}}=\boldsymbol{\mathrm{x_{t}}}-\boldsymbol{\mathrm{x^{n}_{t}}}$, $\delta \boldsymbol{\mathrm{u_{t}}}=\boldsymbol{\mathrm{u_{t}}}-\boldsymbol{\mathrm{u^{n}_{t}}}$.
Given a quadratic control cost, the locally optimal control law will be of the form $\delta\boldsymbol{\mathrm{u_t}}=\boldsymbol{\mathrm{k_t}}+\boldsymbol{\mathrm{K_t}}\delta\boldsymbol{\mathrm{x_t}}$. The underlying optimal control problem can be solved by using Bellman equation
\begin{equation}\label{eq.recursions}
\Psi_\sigma(\delta \boldsymbol{\mathrm{x_t}},\BR{t}) = \min_{\boldsymbol{u}}\{ l(\boldsymbol{\mathrm{x}},\boldsymbol{\mathrm{u}},\BR{t}) + \mathbb{E}[ \Psi_\sigma(\delta \boldsymbol{\mathrm{x_{t+1}}},\BR{t+1})] \}
\end{equation}
where $l$ is the quadratic cost, and by making the following quadratic approximation of the value function $\Psi(\delta \boldsymbol{\mathrm{x_t}},\BR{t}) = \frac{1}{2}\delta \boldsymbol{\mathrm{x_t^T}}\boldsymbol{\mathrm{S_t}}\delta \boldsymbol{\mathrm{x}} + \delta \boldsymbol{\mathrm{x_t^T}}\boldsymbol{\mathrm{s_t}} + s_t$
where $\boldsymbol{\mathrm{S_t}} = \nabla_{\delta x \delta x} \Psi$ and $\boldsymbol{\mathrm{s_t}} = \nabla_{\delta x}\Psi - \boldsymbol{\mathrm{S_t}}\delta \boldsymbol{\mathrm{x_t}}$ are functions of the partial derivatives of the value function.

Using the (time-varying) linear dynamics, the quadratic cost and the quadratic approximation of $\Psi$, and solving for the optimal control, we get
\begin{equation}\delta\boldsymbol{\mathrm{u_t}}=\boldsymbol{\mathrm{k_t}}+\boldsymbol{\mathrm{K_t}}\delta\boldsymbol{\mathrm{x_t}}, \ \ \boldsymbol{\mathrm{k_t}} = -\boldsymbol{\mathrm{H_{t}^{-1}g_t}}, \ \ \textrm{and}\ \ \boldsymbol{\mathrm{K_t}} = -\boldsymbol{\mathrm{H_{t}^{-1}G_t}}
\end{equation}

where $\boldsymbol{\mathrm{H_t}}$, $\boldsymbol{\mathrm{g_t}}$, $\boldsymbol{\mathrm{G_t}}$ are given by
\begin{equation}\label{equation.recursion}
\begin{split}
    & \mathbf{H_t} = \mathbf{R_t} + \mathbf{B_t^T}\mathbf{S_t}\mathbf{B_t} + \sigma \mathbf{B_t^T S_t^T C\Sigma_{t+1} C^T S_tB_t}\\
    & \mathbf{g_t} = \mathbf{r_t} + \mathbf{B_t^T s_t} + \sigma \mathbf{B_t^T S_t^T C\Sigma_{t+1} C^T s_t} \\
    & \mathbf{G_t} = \mathbf{P_t^T} + \mathbf{B_t^T S_t A_t} + \sigma \mathbf{B_t^T S_t^T C\Sigma_{t+1} C^T S_t A_t}
\end{split}
\end{equation}
where
    $ \boldsymbol{\mathrm{A_t}} = \Delta t\frac{\partial\boldsymbol{\mathrm{f}}}{\partial\boldsymbol{\mathrm{x_t}}}$, $\boldsymbol{\mathrm{B_t}} = \Delta t\frac{\partial\boldsymbol{\mathrm{f}}}{\partial\boldsymbol{\mathrm{u_t}}}$ and $\boldsymbol{\mathrm{q_t}}$, $\boldsymbol{\mathrm{r_t}}$, $\boldsymbol{\mathrm{Q_t}}$, $\boldsymbol{\mathrm{R_t}}$ and $\boldsymbol{\mathrm{P_t}}$ are the coefficients of the Taylor expansion of the cost function around the nominal trajectory. The corresponding backward recursions are
\begin{equation}\label{equation.value_recursion}
\mathbf{s_{t}} = \mathbf{q_t} + \mathbf{A_t^T s_{t+1}} + \mathbf{G_t^T k_t} + \mathbf{K_t^T H_t k_t} +  \sigma \mathbf{A_t^T S_{t+1}^T C\Sigma_{t+1} C^T s_{t+1} }
\vspace{-0.4cm}
\end{equation}
\begin{equation}\label{equation.value_recursion2}
\mathbf{S_{t}} = \mathbf{Q_t} + \mathbf{A_t^T S_{t+1} A_t} + \mathbf{K_t^T H_t K_t} + \mathbf{G_t^T K_t} +  \mathbf{K_t^T G_t} + \sigma \mathbf{A_t^T S_{t+1}^T C\Sigma_{t+1} C^T S_{t+1} A_t}
\end{equation}
We note that this Riccati recursion is different from usual iLQR (\citep{Tassa2014})
due to the presence of the covariance $\Sigma$: the locally optimal control law explicitly depends on the noise uncertainty.
\subsection{Curious iLQR: seeking out uncertainties}\label{sec.curious_recursions}
We use Gaussian Process (GP) regression to learn a probabilistic model of the dynamics in order to include the predictive variance
from the model into the risk-sensitive iLQR algorithm. This predictive variance will then capture both model as well as measurement uncertainty. 
Specifically, we set $\BR{x_t} = [\boldsymbol{\mathrm{\theta_t}},\dot{\boldsymbol{\mathrm{\theta_t}}}]$ where $\boldsymbol{\mathrm{\theta_t}}$, $\dot{\boldsymbol{\mathrm{\theta_t}}}$ are joint position and velocity vectors respectively. We let $\BR{u_t}$ denote the vector of commanded torques. After each system rollout, we get a new set of tuples of states and actions $(\BR{x_t},\BR{u_t})$ as inputs and $\BR{\ddot{\theta}_{t+1}}$, joint accelerations at the next time step, as outputs which we add to our dataset $\mathcal{D}$ on which we re-train the probabilistic dynamics model (see Algorithm~\ref{algo:mbrl_loop}). Once trained, the model produces a one step prediction of the joint accelerations 
of the robot as a probability distribution of the form
\begin{equation}\label{eq:prob_dynamics}
p(\BR{\ddot{\theta}_{t+1}}|\BR{x_{t}},\BR{u_{t}}) = \mathcal{N}( \boldsymbol{\mathrm{\ddot{\theta}_{t+1}}} | \boldsymbol{\mathrm{h}}\left( \boldsymbol{\mathrm{x_t}},\boldsymbol{\mathrm{u_t}} \right) \Delta t, \boldsymbol{\Sigma_{t+1}})    
\end{equation}
where $\BR{h}$ is the mean vector and $\BR{\Sigma_{t+1}}$ the covariance matrix of the predictive distribution evaluated at $(\BR{x_t, u_t})$. 
The outputs is the acceleration at the next time step $\boldsymbol{\mathrm{\ddot{\theta}_{t+1}}}$ which is numerically integrated to velocity $    \boldsymbol{\mathrm{\ddot{\theta}_{t+1}}}\Delta t + \boldsymbol{\mathrm{\dot{\theta_t}}} = \boldsymbol{\mathrm{\dot{\theta}_{t+1}}}$  
and position $\boldsymbol{\mathrm{\dot{\theta}_{t+1}}}\Delta t + \boldsymbol{\mathrm{\theta_t}} = \boldsymbol{\mathrm{\theta_{t+1}}}.$ This results in a Gaussian predictive distribution of the system dynamics $\mathbf{f}$
\begin{equation}
    \boldsymbol{\mathrm{x_{t+1}}} \sim \mathcal{N}( \boldsymbol{\mathrm{x_{t+1}}} | \boldsymbol{\mathrm{x_{t}}} + \boldsymbol{\mathrm{h}}\left( \boldsymbol{\mathrm{x_t}},\boldsymbol{\mathrm{u_t}} \right) \Delta t , \boldsymbol{\Sigma_{t+1}})
\end{equation}
It is the covariance matrix $\Sigma_{t+1}$ of this distribution that is incorporated into the Riccati equations from above. Specifically, during each MBRL iteration we optimize a new local feedback policy under the current dynamics model $\mathbf{f}$, via Algorithm~\ref{algo:curious-iLQR}. Each outer loop of the optimization, re-linearizes $\mathbf{f}$ with respect to the current nominal trajectories $\boldsymbol{\mathrm{u^{n}_{t}}}$, $\boldsymbol{\mathrm{x^{n}_{t}}}$ in the backward-pass:
\begin{equation}\label{eq.linarize_dyn_cur_ilqr}
    \delta \boldsymbol{\mathrm{x_{t+1}}} = \boldsymbol{\mathrm{A_{t}}}\delta \boldsymbol{\mathrm{x_{t}}} + \boldsymbol{\mathrm{B_t}}\delta \boldsymbol{\mathrm{u_{t}}} + \boldsymbol{\mathrm{C_{t}}}\omega_{t}
\end{equation}
with $\boldsymbol{\mathrm{A_t}}=\Delta t\frac{\partial\boldsymbol{\mathrm{f}}}{\partial\boldsymbol{\mathrm{{x_t}^n}}}$, $\boldsymbol{\mathrm{B_t}}=\Delta t\frac{\partial\boldsymbol{\mathrm{f}}}{\partial\boldsymbol{\mathrm{{u_t}^n}}}$ and $
    \omega_t \sim \mathcal{N}(\omega_t|0,\BR{\Sigma_{t+1}})$,
where $\BR{A_t}$ and $\BR{B_t}$ are the analytical gradients of the probabilistic model prediction at each time step and $\BR{C_t}$ weights how the uncertainty is propagated trough the system. We utilize the Riccati equations from Section~\ref{sec.risk_sensitive}, Equations~\eqref{equation.recursion} and \eqref{equation.value_recursion}, to optimize a new local feedback policy that utilizes the models predictive covariance $\Sigma_{t+1}$.
During the shooting phase of the algorithm, we integrate the nonlinear model from the GP and,
to guarantee convergence to lower costs, we use a line search approach during the optimization. 
We leverage the risk-seeking capabilities of the optimization by setting $\sigma<0$. 
The algorithm then favors costs with higher variance which is related to exploring regions of the state space with higher uncertainty in the dynamics. As a result, the agent is encouraged to select actions that explore uncertain regions of the dynamic model while still trying to reduce the task specific error. With $\sigma=0$ the agent will ignore any uncertainty in the environment and therefore not explore. This is equivalent to standard iLQR optimization which ignores higher order statistics of the cost function. 
An overview of \curiousilqr  is given in Algorithm \ref{algo:curious-iLQR}.

\section{Illustration: Curious iLQR}\label{section.illustration}
In this section, we want to illustrate the advantages of using the motivation to resolve model uncertainty as an exploration tool. 
The objectives of this section is to give an intuitive example of the effect of our MBRL loop. In the following, and throughout the paper, we will refer to the agent that tries to resolve the uncertainty in its environment as curious and the one that is not following the uncertainty but only optimizes for the task related cost as normal.

\begin{figure}[h]
\vspace{-0.3cm}
\centering
\includegraphics[width=\textwidth]{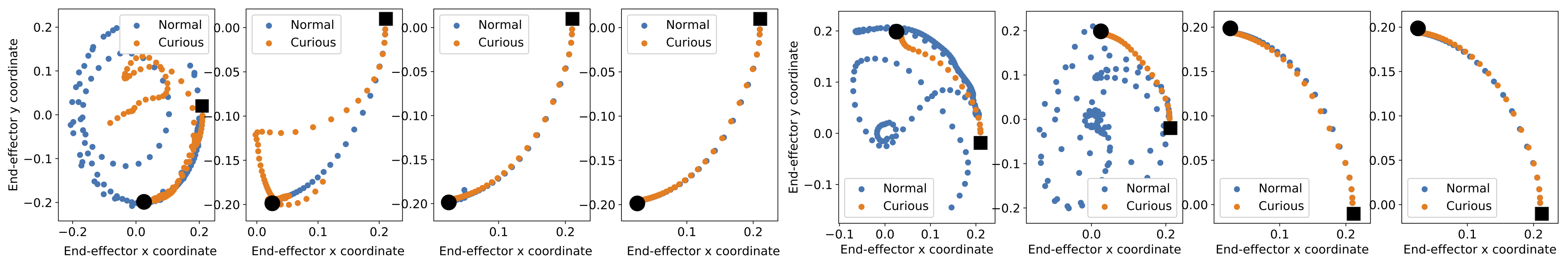}
\caption{End-effector position of curious and normal agent for 4 learning iterations on 2 different targets. The targets are represented by the black dots, the starting position by the black squares.} 
\label{fig:2D_Arm_states}
\vspace{-0.3cm}
\end{figure}
\begin{wrapfigure}{r}{0.4\textwidth}
\vspace{-0.65cm}
\centering
\includegraphics[width=0.4\textwidth]{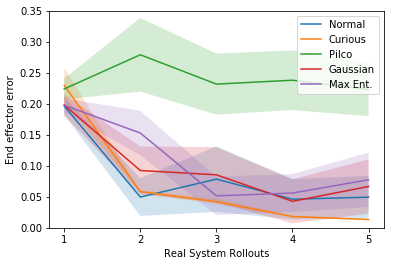}
\caption{\small{Reacher performance /10 trials.}}
\label{fig:comparison_reacher}
\vspace{-0.4cm}
\end{wrapfigure}

The experimental platform is the OpenAI Gym Reacher environment \citep{gym}, a two degrees of freedom arm attached at the center of the scene. The goal of the task is to reach a target placed in the environment. In the experiments presented here, actions were optimized as described in section \ref{section.curious-ilqr}. The probabilistic model was learned with Gaussian Process (GP) regression using the GPy library \citep{gpy2014}. The intuition behind this experiment is that, if an agent is driven to resolve uncertainty in its model, a better model of the system dynamics can be learned and therefore used to optimize a control sequence more reliably. Our hypothesis is that, the model learned by the curious agent is better by the end of learning and therefore we expect it to perform better when using it to solve new reaching tasks.
\begin{figure}[h]
\vspace{-0.3cm}
\centering
\includegraphics[width=0.7\textwidth]{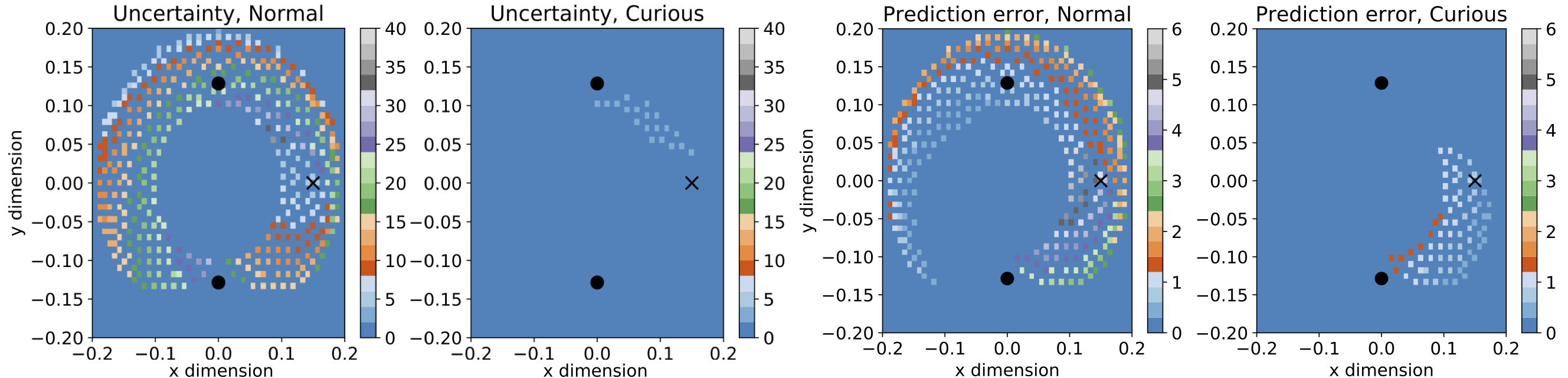}
\caption{\small{The uncertainty and the prediction error in end-effector space after training, for the normal and curious agent. The cross is the initial position. Regions that are not by the arm reachable are shown in blue.}}
\label{fig:2D_Arm_var}
\vspace{-0.4cm}
\end{figure}
In Figure~\ref{fig:2D_Arm_states} we show the resulting end-effector trajectories of 8 consecutive MBRL iterations when optimizing to reach 2 different targets in sequence. We compare the behavior of the curious and normal agent in orange and blue, respectively. The targets are represented by the black dot. The curious agent tries to resolve the uncertainty within the model; the normal agent optimizes only for the task related cost. The normal agent seemingly reaches the first target after the second learning iteration; the curious agent only manages to reach the target during the third iteration. Interestingly, the exploration of the curious agent leverages the arm to reach the second target immediately and continues to reach it consistently thereafter. Figure \ref{fig:2D_Arm_var} confirms the intuition that the curious agent has learned a better model than the normal agent. The figure shows the uncertainty and the prediction error (in end-effector space) of the model learned by the normal and the curious agent respectively. With curiosity, the learned model has overall lower uncertainty and prediction error values over the whole state space. We also compare our MBRL loop via \curiousilqr  optimization to: normal iLQR, a random exploration controller that adds Gaussian noise to the actions with mean $0$ and variance $0.2$, a maximum entropy exploration behaviour following the approach proposed in \citep{levine2016end} and PILCO \citep{deisenroth2011pilco}, in Figure \ref{fig:comparison_reacher}. For these experiments, we initialize the model with only two data points collected randomly during motor babbling. We report the mean and the standard deviation across 10 trials, where each trial starts from a different initial joint configuration and is initialized with a different initial torque trajectory for optimization. In this scenario, with a very poor initial model quality, PILCO could not perform comparably to our MBRL loop. MBRL via \curiousilqr  outperforms all the other approaches. Furthermore it converges to solutions more reliably, as the variance between trials is lowest. 
\section{Experiments on high-dimensional problems}\label{section.experiments}
Finally, the goal of this work is to learn motor skills on a torque-controlled manipulator. Our experimental platform is the Sawyer robot from Rethink Robotics \citep{sawyer}, a 7 degrees of freedom manipulator. We start with experiments performed in the PyBullet physics simulator \citep{pybullet}. In the next Section, we present results on the Sawyer robot arm hardware. Previous work such as \citep{Farshidian} and \citep{Ponton}, which use risk-sensitive control variations of iLQR, primarily deal with simplified, low dimensional problems. Our experiments are conducted on a 7 degree of freedom robot, and the higher dimensional system adds some complexities to the approach: the gradients in Section \ref{sec.risk_sensitive} of the value function (Equations \eqref{equation.value_recursion}, \eqref{equation.value_recursion2}) tend  to suffer from numerical ill-conditioning in high-dimensions. We account for this issue with Tikhonov regularization: before inversion for calculating the optimal control we add a diagonal matrix to $\boldsymbol{\mathrm{H_k}}$ from Equation \eqref{equation.recursion}. The regularization parameter and the line search parameter $\alpha$ are adapted following the Levenberg Marquardt heuristic \citep{Tassa2014}. 

The goal of these experiments is to reach a desired target joint configuration $\theta$. We show results for dynamics learned with GP regression (GPR), as well as initial results on ensemble of probabilistic neural networks (EPNN) following the approach presented in \citep{lakshminarayanan2017simple}. When using GPs, a separate GP is trained for each output dimension. 

We perform two sets of experiments, both in simulation and on hardware, to analyze the effect of using curiosity. Specifically, we believe that curiosity helps to find better solutions faster, because it guides exploration within the MBRL loop. Intuitively, curiosity helps to observe more diverse data samples during each rollout such that the model learns more about the dynamics. 

We start with evaluation in simulation. Throughout all of the simulation experiments the optimization horizon was 150 time steps long at a sampling rate of $240$ Hz. Motor babbling was performed at the beginning for 0.5$s$ by commanding random torques in each joint.
\subsection{Reaching task from scratch}\label{exp.1}
During the first set of experiments, we compare the performance when learning to reach a given target configuration from scratch. We compare our MBRL loop, as before, when using our \curiousilqr  optimization, regular iLQR, a random exploration controller and a maximum entropy exploration behaviour as described previously. PILCO was not able to learn the reaching movement on the 7-DoF manipulator, so we exclude the results from the analysis. We perform this experiments for each kind of controller 5 times. Each run slightly perturbs the initial joint positions, and uses a random initial torque trajectories for the optimizer. For a given target joint configuration, $5$ iterations of optimizing the trajectory, running the trajectory on the system and updating the dynamics model, were performed. We perform this experiment for $3$ different target joint configurations. The following results are averaged across the $5x3$ runs ($5$ runs per target).
\begin{figure}[ht]
\vspace{-0.7cm}
\centering
\includegraphics[width=0.8\textwidth]{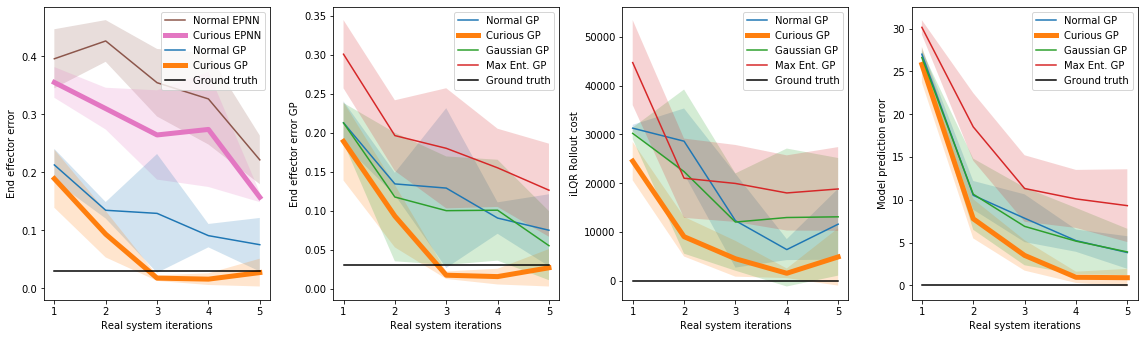}
\caption{\small{Distance in end effector space for EPNN vs. GP in m (1). Distance in end effector space in m (2), iLQR rollout cost (3) and model prediction error (4) with the GP model, compared to our baselines.}}
\label{fig:model_performance}
\vspace{-0.66cm}
\end{figure}
The left most plot in Figure~\ref{fig:model_performance} compares performance of \curiousilqr when using EPNN vs GPR for dynamics model learning, with and without curiosity. Our  analysis shows that MBRL via \curiousilqr improves performance over regular iLQR, for both model architectures. While the EPNN is more promising in scaling our approach, it currently requires more data to train. For this reason we will focus on the GP model for the remainder of our experimental section. In the 2nd to 4th plot of Figure~\ref{fig:model_performance}, we compare the performance of \curiousilqr against the above mentioned baselines for exploration during policy optimization, when using GPR for model learning. We compare the methods with respect to 3 metrics: final Euclidean end-effector distance (plot 2), iLQR cost (plot 3) and the predictive performance of the model on each rollout (plot 4). We can consistently see that, on average, MBRL via \curiousilqr  outperforms the other approaches: the error/cost is smaller and the solutions are more consistent across trials as the standard deviation is lower. This shows that curiosity can lead to faster learning of a new task, when learning from scratch.
The results on the predictive performance of the model suggest that the quality of the model learned via \curiousilqr  might be better in terms of generalization. In the next section we present results that investigate this assumption.

\subsection{Optimizing towards new targets after model learning}\label{exp.2}
\begin{figure}[h]
\vspace{-0.5cm}
\centering
\includegraphics[width=0.77\textwidth]{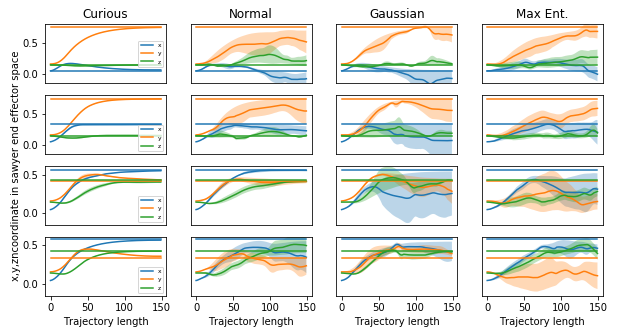}
\caption{\small{Optimizing to reach new targets with regular iLQR after models were learned. $4$ different targets (one per row) are evaluated and the final end-effector trajectories presented. Constant lines are targets for x/y/z.}}
\label{fig:reaching_new_targets}
\vspace{-0.40cm}
\end{figure}
To confirm the hypothesis that the models learned by MBRL with curious iLQR generalize better, because they have explored the state space better, we decided to evaluate the learned dynamics models on a second set of experiments in which the robot tries to reach new, unseen targets. In this experiment we take the GP models learned during experiment 1 in Section~\ref{exp.1} and use them to optimize trajectories to reach new targets that were not seen during training of the model. The results are shown in Figure~\ref{fig:reaching_new_targets}, where four randomly chosen targets were set and the trajectory was optimized with regular iLQR. Note, that here we use regular iLQR to optimize for the trajectory so that we can better compare the models learned with/without curiosity in the previous set of experiments. Figure \ref{fig:reaching_new_targets} shows the trajectory in end effector space for each coordinate dimension, together with the target end effector position as a solid horizontal line. 
The results are averaged across $5$ trials. The trials correspond to using one of the $5$ dynamics models at the end of Experiment 1 in Section~\ref{exp.1}. For each trial, the initial torque trajectory was initialized randomly, and the initial joint configuration slightly perturbed. The mean and the standard deviation of the optimized trajectories are computed across the $5$ models learned via MBRL with curious iLQR (first col), MBRL with normal iLQR (second col), iLQR with random exploration (third col) and iLQR with maximum entropy exploration bonus (fourth col.). We see that MBRL with curious iLQR results in a model that performs better when presented with a new target. The new targets are reached more reliably and precisely. 
\section{Real hardware experiments}\label{seq.real_hardware}
\begin{minipage}{0.6\textwidth}
\vspace{-0.7cm}
\begin{table}[H]
 \scalebox{0.6}{
\begin{tabular}{|c|l|l|l|l|l|l|l|l|}
\hline
\textbf{Target}        & \multicolumn{8}{c|}{\textbf{Distance to Target in m (Learning Iteration)}}                                        \\ \hline
\multicolumn{1}{|l|}{} & \multicolumn{4}{c|}{\textbf{Curious}}                 & \multicolumn{4}{c|}{\textbf{Normal}}                 \\ \hline
\textbf{1}             & 0.05 (6) & 0.09 (2) & 0.09 (3) & \textbf{0.07 (3.67)} & 0.37 (8) & 0.08 (2) & 0.18 (8) & \textbf{0.21 (7.0)} \\ \hline
\textbf{2}             & 0.05 (3) & 0.09 (4) & 0.09 (4) & \textbf{0.07 (3.67)} & 0.20 (8) & 0.08 (3) & 0.09 (5) & \textbf{0.12 (5.3)} \\ \hline
\textbf{3}             & 0.09 (6) & 0.09 (4) & 0.09 (3) & \textbf{0.09 (4.33)} & 0.17 (8) & 0.16 (8) & 0.11 (8) & \textbf{0.15 (8.0)} \\ \hline
\textbf{4}             & 0.04 (2) & 0.07 (2) & 0.07 (2) & \textbf{0.06 (2.33)} & 0.04 (3) & 0.08 (3) & 0.05 (3) & \textbf{0.06 (3.0)} \\ \hline
\multicolumn{1}{|l|}{} &          &          &          & \textbf{0.07 (3.5)}  &          &          &          & \textbf{0.14 (5.9)} \\ \hline
\end{tabular}
}
\caption{\small{Results on a reaching task. Each task (target) was repeated three times. The mean values are reported in bold font.}}\label{table.sawyer}
\end{table}
\vspace{-0.7cm}
\end{minipage}
\hfill
\begin{minipage}{0.3\textwidth}
\vspace{-0.7cm}
\begin{table}[H]
\centering
\scalebox{0.6}{
\begin{tabular}{|l|l|l|}
\hline
       & \multicolumn{2}{l|}{\textbf{\begin{tabular}[c]{@{}l@{}}Reaching Precision (m)\end{tabular}}} \\ \hline
Target & \textbf{Curious}                                & \textbf{Normal}                               \\ \hline
1      & 0.20                                            & 0.67                                          \\ \hline
2      & 0.26                                            & 0.61                                          \\ \hline
3      & 0.25                                            & 1.06                                          \\ \hline
4      & 0.24                                            & 0.67                                          \\ \hline
5      & 0.37                                            & 0.49                                          \\ \hline
       & \textbf{0.26}                                   & \textbf{0.7}                                  \\ \hline
\end{tabular}
}
\caption{\small{Reaching a new target not seen during training.}}\label{table.new.target.sawyer}
\end{table}
\vspace{-0.7cm}
\end{minipage}

The experimental platform for our hardware experiments is the Sawyer Robot \citep{sawyer}. The purpose of the experiments was to demonstrate the applicability and the benefits of our algorithm on real hardware. We perform reaching experiments for 4 different target locations. Each experiment is started from scratch with no prior data, and the number of hardware experiments needed to reach the target are compared. The results are summarized in Table~\ref{table.sawyer} and show the number of learning iterations needed in order to reach the target together with the precision in end-effector space. If the target was reached with a precision of below 10 cm, we would consider the task as achieved; if the target was not reached after the 8th learning iteration we would stop the experiment and consider the last end-effector position. We decided to terminate our experiments after the eight iteration as running the experiment on hardware was a lengthy process, as the GP training and the rollout would happen iteratively and GP training time increases with growing amount of data. Also, the reaching precision that we were able to achieve on hardware was significantly lower, compared to the simulation experiments. 
\begin{wrapfigure}{l}{0.45\textwidth}
\vspace{-0.2cm}
\centering
  \begin{subfigure}[b]{0.22\textwidth}
  \centering
    \includegraphics[width=\textwidth]{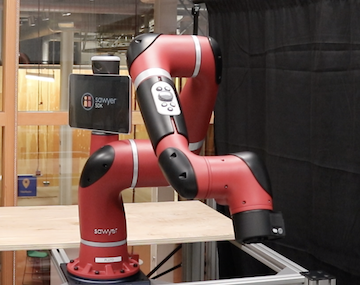}
    \caption{\small{start configuration}}
    \label{fig:f1}
  \end{subfigure}
  \hfill
  \begin{subfigure}[b]{0.22\textwidth}
  \centering
    \includegraphics[width=\textwidth]{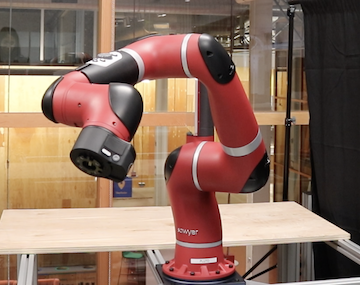}
    \caption{\small{target configuration}}
    \label{fig:f2}
  \end{subfigure}
  \caption{Joint configuration of Sawyer.}
  \vspace{-0.4cm}
\end{wrapfigure}
We believe this is due to the data collected from the Sawyer robot, as we could only control the robot at $100Hz$ which introduces inaccuracies when reading the effects of the sent torque command. We repeated each experiment three times to demonstrate the repeatability of our method as we expected measurement noise to affect solutions.
From the table we can see that MBRL with curious iLQR would reach a target on average after 3.5 iterations with an average precision of 7 cm, compared to MBRL with regular iLQR that needed 5.9 iterations (often not ever reaching the target after eight iterations with the desired precision), with a precision of 14cm on average.
As in simulation, similar to Experiment~\ref{exp.2} we wanted to evaluate the quality of the learned models on new target positions. The results are summarized in Table~\ref{table.new.target.sawyer} and are similar to what we observe in simulation: the models learned with curiosity, when used to optimize for new targets, can achieve higher precision than when using the models learned without curiosity.
\section{Conclusion and future work}\label{section.conclusion}
In this work, we presented a model-based reinforcement learning algorithm that uses an optimal control framework to trade-off between optimizing for a task specific cost and exploring around a locally optimal trajectory. Our algorithm explicitly encourages actions that seek out uncertainties in our model by incorporating them into the cost. By doing so, we are able to learn a model of the dynamics that achieves the task faster than MBRL with standard iLQR, and also transfers well to other tasks. We present experiments on a Sawyer robot in simulation and on hardware. In both sets of experiments, MBRL with \curiousilqr  (our approach) not only learns to achieve the specified task faster, but also generalizes to new tasks and initial conditions. All this points towards the conclusion that resolving dynamics uncertainty during model-based reinforcement learning is indeed a powerful tool.
As \citep{loewenstein1994psychology} states, curiosity is a superficial affection: it can arise, diverge and end promptly. We were able to observe similar behaviour in our experiments as well, as can be seen in Figure \ref{fig:model_performance}: towards the end of learning, the exploration signal around the trajectory decreases and the robot would explore, deviate from the task slightly, before going back to exploiting once it is fairly certain about the dynamics. In the future, we would like to explore this direction by considering how to maintain exploration strategies. This could be helpful if the robot is still certain about a task, even though the environment or task has changed.

\clearpage
\acknowledgments{The authors would like to thank Stefan Schaal for his advise throughout the project. The authors thank the International Max Planck Research School for Intelligent Systems (IMPRS-IS) for supporting Sarah Bechtle. Part  of  this  work  was  supported  by  the  Max-Planck  Society,  New  York University, the   European   Unions   Horizon   2020   research   and   innovation program  (grant  agreement  780684  and  European  Research  Councils  grant 637935) and a Google Faculty Research Award.}


\bibliography{main}  

\appendix
\section{Background: Uncertainty in Optimal Control}\label{section.appendix.background}

In this appendix we review the fundamentals on which our work is built on. An approach enabling the inclusion of higher order statistics in the performance measure while keeping computations tractable, at least in the linear case, is to use exponential costs, as introduced by Jacobson \citep{Jacobson1973}. 
\citep{Farshidian} extended this work by deriving an iterative algorithm for continuous-time stochastic nonlinear optimal control problems called iterative Linear Exponential-Quadratic Optimal Control under Gaussian Process Noise (iLEG). Ponton et al. \citep{Ponton} extended the work from \citep{Farshidian} to cases where not only process noise is present but also measurement noise has to be taken into consideration.
Next, we briefly present the details of the risk-sensitive iLQR algorithm, following \citep{Farshidian}, \citep{Ponton} and \citep{Jacobson1973}.
\subsection{Risk-sensitive iLQR}\label{sec.appendix.risk_sensitive}
To include stochastic processes when optimizing a trajectory, it is necessary to consider a nonlinear optimal control problem where the system dynamics are defined by the following stochastic differential equation
\begin{equation}\label{eq.appendix.dynamics}
    \boldsymbol{\mathrm{x_{k+1}}}=\boldsymbol{\mathrm{x_k}}+\boldsymbol{\mathrm{f}}\left( \boldsymbol{\mathrm{x_k}},\boldsymbol{\mathrm{u_k}} \right) \Delta t +\boldsymbol{\mathrm{g}}\left( \boldsymbol{\mathrm{x_k},\mathrm{u_k}}\right) \Delta \omega 
\end{equation}
where $\boldsymbol{\mathrm{f}}$ represents the dynamics of the system and $\boldsymbol{\mathrm{g}}$ the stochasticity of the problem. $\Delta \omega$ is a Brownian motion with zero mean and covariance $(\boldsymbol{\Sigma} \cdot \Delta t)$. Following the idea of \citep{Jacobson1973} to include higher order momenta of the cost function, the objective function takes the form of an exponential transformation of the performance criteria $J$:
\begin{equation}\label{eq.appendix.cost}
  J=\min _{ \pi  }{\mathbb{E}\left\{\mathrm{exp}[\sigma \mathcal{J}  \left( \pi  \right)  \right]\} }\end{equation}
  where $\mathcal{J}(\pi)$ is the performance index, which is a random variable, and a functional of the policy $\pi$. $\mathbb{E}$ $\mathbb{E}$ is the expected value of $\mathcal{J}$ over stochastic trajectories induced by the policy $\pi$.
  $\sigma \in \mathbb{R}$ accounts for the sensitivity of the cost to higher order moments (variance, skewness, etc). 
  Notably from \citep{Farshidian}, the cost is
  \begin{equation}
\frac{1}{\sigma} \log(J)  = \mathbb{E}(\mathcal{J}^*)+ \frac{\sigma}{2} \mathrm{var}(\mathcal{J}^*) + \frac{\sigma^2}{6} \mathrm{sk}(\mathcal{J}^*) + \cdots
\end{equation}
where $\mathrm{var}$ and $\mathrm{sk}$ stand for variance and skewness and $\mathcal{J}^*$ is the optimal task cost.
  When $\sigma>0$ the optimal control will be risk-averse, favoring low costs with low variance but when $\sigma<0$ the optimal control will be risk-seeking, favoring low costs with high variance. 
  $\sigma =0$ reduces to the standard, risk-neutral, optimal control problem.
  We will exploit this property to create policies that explore regions with high uncertainty in the next sections.
   
Let the performance criteria $\mathcal{J}$ be 
\begin{equation}
    \mathcal{J}={ \Phi  }_{ f }\left( \boldsymbol{\mathrm{{ x }_{ { T }}}} \right) + \BR{\sum_{{k=0}}^{T}}  L(\boldsymbol{\mathrm{ x }_\mathrm{ k } },\boldsymbol{\mathrm{ u }_\mathrm{ k } })
\end{equation}
where
\begin{equation}
    L(\boldsymbol{\mathrm{{ x }_{ k }}},\boldsymbol{\mathrm{{ u }_{ k }}}) = \Phi(\boldsymbol{\mathrm{x_{k}}}) + \frac { 1 }{ 2 } \boldsymbol{\mathrm{u_{ k }^{ T }}}\boldsymbol{\mathrm{R}}(\boldsymbol{\mathrm{x_{k}}})\boldsymbol{\mathrm{u_{k}}} + \boldsymbol{\mathrm{u_{ k }^{ T }}}\boldsymbol{\mathrm{r}}(\boldsymbol{\mathrm{x_k}})
\end{equation}
$L$ is composed of the state cost $\Phi(\boldsymbol{\mathrm{x_{k}}})$ and a quadratic control cost. ${ \Phi  }_{ f }\left( \BR{x_T} \right)$ is the cost at the final state. 
\subsection{Algorithm derivation}
The algorithm begins with a nominal state and control input trajectory $\boldsymbol{\mathrm{x^{n}}}$ and $\boldsymbol{\mathrm{u^{n}}}$. The dynamics are linearized and the cost is quadratized along $\boldsymbol{\mathrm{u^{n}_{k}}}$, $\boldsymbol{\mathrm{x^{n}_{k}}}$ in terms of state and control deviations
$\delta \boldsymbol{\mathrm{x_{k}}}=\boldsymbol{\mathrm{x_{t}}}-\boldsymbol{\mathrm{x^{n}_{k}}}$, $\delta \boldsymbol{\mathrm{u_{k}}}=\boldsymbol{\mathrm{u_{k}}}-\boldsymbol{\mathrm{u^{n}_{k}}}$ leading to the linear dynamics approximation:
\begin{equation}\label{eq.appendix.linarize_dyn}
    \delta \boldsymbol{\mathrm{x_{k+1}}} = \boldsymbol{\mathrm{A_{k}}}\delta \boldsymbol{\mathrm{x_{k}}} + \boldsymbol{\mathrm{B_k}}\delta \boldsymbol{\mathrm{u_{k}}} + \boldsymbol{\mathrm{C_{k}}}\omega_{k}
\end{equation}
where
    $ \boldsymbol{\mathrm{A_k}} = \Delta t\frac{\partial\boldsymbol{\mathrm{f}}}{\partial\boldsymbol{\mathrm{x_k}}}$ and $\boldsymbol{\mathrm{B_k}} = \Delta t\frac{\partial\boldsymbol{\mathrm{f}}}{\partial\boldsymbol{\mathrm{u_k}}}$

$\boldsymbol{\mathrm{C_k}}$ represents how the uncertainty propagates through the system. And the quadratic approximation of the cost
\begin{equation}
\begin{split}
    \tilde{l}(\boldsymbol{\mathrm{x}},\boldsymbol{\mathrm{u}},\BR{k}) = q_{\BR{k}} + \boldsymbol{\mathrm{q_k^T}}\delta\boldsymbol{\mathrm{x_k}}+\boldsymbol{\mathrm{r_k^T}}\delta\boldsymbol{\mathrm{u_k}} +  \frac{1}{2}\delta\boldsymbol{\mathrm{x^T}}\boldsymbol{\mathrm{Q_k}}\delta\boldsymbol{\mathrm{x_k}} +  \delta\boldsymbol{\mathrm{x^T}}\boldsymbol{\mathrm{P_k}}\delta\boldsymbol{\mathrm{u_k}} + \frac{1}{2}\delta\boldsymbol{\mathrm{u^T}}\boldsymbol{\mathrm{R_k}}\delta\boldsymbol{\mathrm{u_k}}
\end{split}
\end{equation}
with final cost 
\begin{equation}
    \tilde{l}(\BR{x}, \BR{N})=q_{\BR{N}} + \boldsymbol{\mathrm{q_N^T}}\delta\boldsymbol{\mathrm{x_k}} + \frac{1}{2} \delta\boldsymbol{\mathrm{x_k^T}}\boldsymbol{\mathrm{Q_N}}\delta\boldsymbol{\mathrm{x_k}}
\end{equation}
where $q_{\BR{k}}$ $\boldsymbol{\mathrm{q_k}}$, $\boldsymbol{\mathrm{r_k}}$, $\boldsymbol{\mathrm{Q_k}}$, $\boldsymbol{\mathrm{R_k}}$ and $\boldsymbol{\mathrm{P_k}}$ are the coefficients of the Taylor expansion of the cost function around the nominal trajectory. Given the quadratic control cost, the locally optimal control law will be of the form $\delta\boldsymbol{\mathrm{u_k}}=\boldsymbol{\mathrm{k_k}}+\boldsymbol{\mathrm{K_k}}\delta\boldsymbol{\mathrm{x_k}}$ and the underlying optimal control problem can be solved by using Bellman equation
\begin{equation}\label{eq.appendix.recursions}
\Psi_\sigma(\delta \boldsymbol{\mathrm{x_k}},\BR{k}) = \min_{\boldsymbol{u}}\{ \tilde{l}(\boldsymbol{\mathrm{x}},\boldsymbol{\mathrm{u}},\BR{k}) + \mathbb{E}[ \Psi_\sigma(\delta \boldsymbol{\mathrm{x_{k+1}}},\BR{k+1})] \}
\end{equation}
which becomes is:
\begin{equation}
\begin{split}
 \frac{1}{2}\delta \boldsymbol{\mathrm{x_k^T}}\boldsymbol{\mathrm{S_{k}}}\delta \boldsymbol{\mathrm{x}} + \delta \boldsymbol{\mathrm{x_k^T}}\boldsymbol{\mathrm{s_{k}}} + s_{k} = \min_{u_t} \{ q_{k} + \boldsymbol{\mathrm{q_k^T}}\delta\boldsymbol{\mathrm{x_k}}+\boldsymbol{\mathrm{r_k^T}}\delta\boldsymbol{\mathrm{u_k}} + \frac{1}{2}\delta\boldsymbol{\mathrm{x^T}}\boldsymbol{\mathrm{Q_k}}\delta\boldsymbol{\mathrm{x_k}} + \delta\boldsymbol{\mathrm{x^T}}\boldsymbol{\mathrm{P_k}}\delta\boldsymbol{\mathrm{u_k}} + \\ + \frac{1}{2}\delta\boldsymbol{\mathrm{u^T}}\boldsymbol{\mathrm{R_k}}\delta\boldsymbol{\mathrm{u_k}} + \frac{1}{2}(\boldsymbol{\mathrm{A_{k}}}\delta \boldsymbol{\mathrm{x_{k}}} + \boldsymbol{\mathrm{B_k}}\delta \boldsymbol{\mathrm{u_{k}}})^T \boldsymbol{\mathrm{S_{k+1}}} (\boldsymbol{\mathrm{A_{k}}}\delta \boldsymbol{\mathrm{x_{k}}} + \boldsymbol{\mathrm{B_k}}\delta \boldsymbol{\mathrm{u_{k}}}) + (\boldsymbol{\mathrm{A_{k}}}\delta \boldsymbol{\mathrm{x_{k}}} + \boldsymbol{\mathrm{B_k}}\delta \boldsymbol{\mathrm{u_{k}}})^T \boldsymbol{\mathrm{s_{k+1}}}+\\
 + \frac{\sigma}{2}(\mathrm{\boldsymbol{S_{k+1}}}(\boldsymbol{\mathrm{A_{k}}}\delta \boldsymbol{\mathrm{x_{k}}} + \boldsymbol{\mathrm{B_k}}\delta \boldsymbol{\mathrm{u_{k}}})+\boldsymbol{\mathrm{s_{k+1}}})^T \boldsymbol{\mathrm{S_{k+1} C\Sigma C^T}} (\mathrm{\boldsymbol{S_{k+1}}}(\boldsymbol{\mathrm{A_{k}}}\delta \boldsymbol{\mathrm{x_{k}}} + \boldsymbol{\mathrm{B_k}}\delta \boldsymbol{\mathrm{u_{k}}})+\boldsymbol{\mathrm{s_{k+1}}}) +\\+ \frac{1}{2}Tr(\boldsymbol{\mathrm{S_{k+1} C\Sigma C^T}})\}
\end{split}
\end{equation}
and making the following quadratic approximation of the value function $\Psi$:
\begin{equation}
    \Psi(\delta \boldsymbol{\mathrm{x_k}},\BR{k}) = \frac{1}{2}\delta \boldsymbol{\mathrm{x_k^T}}\boldsymbol{\mathrm{S_k}}\delta \boldsymbol{\mathrm{x}} + \delta \boldsymbol{\mathrm{x_k^T}}\boldsymbol{\mathrm{s_k}} + s_k
\end{equation}
where $\boldsymbol{\mathrm{S_k}} = \nabla_{\delta x \delta x} \Psi$ and $\boldsymbol{\mathrm{s_k}} = \nabla_{\delta x}\Psi - \boldsymbol{\mathrm{S_k}}\delta \boldsymbol{\mathrm{x_k}}$ are functions of the partial derivatives of the value function.
Using the linear dynamics, the quadratic cost and the quadratic approximation of $\Psi$, and solving for the optimal control, we get
\begin{equation}
\begin{split}
        & \delta\boldsymbol{\mathrm{u_k}}=\boldsymbol{\mathrm{k_k}}+\boldsymbol{\mathrm{K_k}}\delta\boldsymbol{\mathrm{x_k}} \\
        & \boldsymbol{\mathrm{k_k}} = -\boldsymbol{\mathrm{H_{k}^{-1}g_k}}\\
        & \boldsymbol{\mathrm{K_k}} = -\boldsymbol{\mathrm{H_{k}^{-1}G_k}}
\end{split}
\end{equation}
where $\boldsymbol{\mathrm{H_k}}$, $\boldsymbol{\mathrm{g_k}}$, $\boldsymbol{\mathrm{G_k}}$ are given by
\begin{equation}\label{equation.appendix.recursion}
\begin{split}
    & \mathbf{H_k} = \mathbf{R_k} + \mathbf{B_k^T}\mathbf{S_k}\mathbf{B_k} + \sigma \mathbf{B_k^T S^T C\Sigma C^T S_kB_k}\\
    & \mathbf{g_k} = \mathbf{r_k} + \mathbf{B_k^T s_k} + \sigma \mathbf{B_k^T S_k^T C\Sigma C^T s_k} \\
    & \mathbf{G_k} = \mathbf{P_k^T} + \mathbf{B_k^T S_k A_k} + \sigma \mathbf{B_k^T S_k^T C\Sigma C^T S_k A_k}
\end{split}
\end{equation}
The corresponding backward recursions are
\begin{equation}\label{equation.appendix.value_recursion}
\begin{split}
\mathbf{s_{k}} = \mathbf{q_k} + \mathbf{A_k^T s_{k+1}} + \mathbf{G_k^T k_k} + \mathbf{K_k^T H_k k_k} +  \sigma \mathbf{A_k^T S_{k+1}^T C\Sigma C^T s_{k+1} }
\end{split}
\end{equation}
\begin{equation}\label{equation.appendix.value_recursion2}
\begin{split}
\mathbf{S_{k}} = \mathbf{Q_k} + \mathbf{A_k^T S_{k+1} A_k} + \mathbf{K_k^T H_k K_k} + \mathbf{G_k^T K_k} +  \mathbf{K_k^T G_k} + \sigma \mathbf{A_k^T S_{k+1}^T C\Sigma C^T S_{k+1} A_k}
\end{split}
\end{equation}
With $\sigma=0$ the recursions revert to the usual Ricatti recursions for iLQR \citep{Tassa2014}.

\section{Details for Experiments}
Throughout the experiments we chose $\sigma=-0.05$ for the curious robot, and $\sigma=0.0$ for the normal robot. The iLQR position error weight $Q_{pos}=5.0$ the velocity weight $Q_{vel}=0.1$ and the torque error weight $R=10^{-7}$. The regularization parameter $\lambda$ was initialized with $1$, the scaling factor for $\lambda$ was $10$ and the maximum $\lambda$ value allowed was $1000$. 

\end{document}